\documentclass[sigconf]{acmart}

\usepackage{array}
\newcolumntype{L}[1]{>{\raggedright\let\newline\\\arraybackslash\hspace{0pt}}m{#1}}
\newcolumntype{C}[1]{>{\centering\let\newline\\\arraybackslash\hspace{0pt}}m{#1}}
\newcolumntype{R}[1]{>{\raggedleft\let\newline\\\arraybackslash\hspace{0pt}}m{#1}}
\usepackage{booktabs}
\usepackage{tabularx}
\usepackage{xcolor,colortbl}
\usepackage{floatflt}

\definecolor{White}{rgb}{1,1,1}
\definecolor{Gray}{gray}{0.9}

\usepackage{soul}

\usepackage{hyperref}
\hypersetup{colorlinks=true,
citecolor=blue,
linkcolor=black,
urlcolor=black}

\setcopyright{rightsretained}

\acmDOI{10.1145/3071178/3071215}

\acmISBN{ 978-1-4503-4920-8/17/07}

\acmConference[GECCO '17]{the Genetic and Evolutionary Computation Conference 2017}{July 15--19, 2017}{Berlin, Germany}
\acmYear{2017}
\copyrightyear{2017}

\begin{document}
\title{Ensemble representation learning: an analysis of fitness and survival for wrapper-based genetic programming methods}

\author{William La Cava}
\authornote{Corresponding Author}
\orcid{1234-5678-9012}
\affiliation{%
  \institution{University of Pennsylvania}
  \streetaddress{3700 Hamilton Walk}
  \city{Philadelphia} 
  \state{PA} 
  \postcode{19104}
}
\email{lacava@upenn.edu}

\author{Jason H. Moore}
\affiliation{%
  \institution{University of Pennsylvania}
  \streetaddress{3700 Hamilton Walk}
  \city{Philadelphia} 
  \state{PA} 
  \postcode{19104}
}
\email{jhmoore@upenn.edu}

\begin{abstract}
Recently we proposed a general, ensemble-based feature engineering wrapper (FEW) that was paired with a number of machine learning methods to solve regression problems. Here, we adapt FEW for supervised classification and perform a thorough analysis of fitness and survival methods within this framework. Our tests demonstrate that two fitness metrics, one introduced as an adaptation of the silhouette score, outperform the more commonly used Fisher criterion. We analyze survival methods and demonstrate that $\epsilon$-lexicase survival works best across our test problems, followed by random survival which outperforms both tournament and deterministic crowding. We conduct a benchmark comparison to several classification methods using a large set of problems and show that FEW can improve the best classifier performance in several cases. We show that FEW generates consistent, meaningful features for a biomedical problem with different ML pairings.
\end{abstract}

%


\maketitle

\section{Introduction}
%


When traditional genetic programming (GP) is applied to classification and/or regression, individual programs assume the roles of feature selection, transformation, and model prediction, and are evaluated for their ability to make accurate estimations and/or predictions. The flexibility of evolving the structure and parameters of a model comes with a heavy computational cost that can be mitigated if one instead uses a fast (e.g. polynomial-time) machine learning (ML) method to optimize the parameters of a GP model with respect to an objective function (for example, least squares error minimization with linear regression). With this in mind, many variants of GP have been proposed that embed linear regression and/or local search in each program, leading to better models~\cite{iba_genetic_1994,kommenda_effects_2013,arnaldo_multiple_2014,la_cava_genetic_2015}. The high-level takeaway from the success of methods that hybridize GP is that it is best to focus the computational effort of GP on the parts of the modeling process that are known to be NP-hard, namely the tasks of feature selection~\cite{foster_variable_2015} and construction~\cite{krawiec_genetic_2002}. 

The task of feature construction, also known as feature engineering or representation learning, is well-motivated since the central factor affecting the quality of a model derived from ML is the ability of the data representation to facilitate learning~\cite{bengio_representation_2013}. This paper focuses on the supervised classification task, for which the goal is to find a mapping ${\hat{y}(\mathbf{x}): \mathbb{R}^d \rightarrow \mathcal{Y}}$ that associates the vector of attributes $\mathbf{x} \in \mathbb{R}^d$ with $k$ class labels from the set $\mathcal{Y} = \{1\;\dots\;k\}$ using $N$ paired examples $\mathcal{T} = \{(\mathbf{x}_i,y_i)\}_{i = 1}^{N}$. The goal of feature engineering is to find a new representation of $\mathbf{x}$ via a $P$-dimensional feature mapping $\mathbf{\Phi}(\mathbf{x}): \mathbb{R}^d \rightarrow \mathbb{R}^P$, such that a classifier ${\hat{y}(\mathbf{\Phi}(\mathbf{x})): \mathbb{R}^P \rightarrow \mathcal{Y}}$ more accurately classifies samples than $\hat{y}(\mathbf{x})$.


GP-based approaches to representation learning include evolving single features for decision trees (DT)~\cite{muharram_evolutionary_2005}, or coupling ML with each program~\cite{krawiec_genetic_2002, silva_multiclass_2015,zegklitz_symbolic_2017}. Recent work~\cite{de_melo_kaizen_2014,arnaldo_building_2015} has advocated what we refer to as an ``ensemble" approach which treats the entire GP population as $\mathbf{\Phi}(\mathbf{x})$, with each program representing a transformation of the form $\phi(\mathbf{x}): \mathbb{R}^d \rightarrow \mathbb{R}$. These proposed methods feed the population output $\mathbf{\Phi}(\mathbf{x}) = [\phi_1(\mathbf{x})\;\dots\;\phi_P(\mathbf{x})]$ into a linear regression model to make predictions. 

The ML-specific nature of these previous approaches motivates our development of the more general feature engineering wrapper (FEW) method\footnote{Available from \url{https://lacava.github.io/few} and via the Python Package Index: \url{https://pypi.python.org/pypi/FEW}}, which is a wrapper-based ensemble method of feature engineering with GP~\cite{la_cava_general_2017}. Unlike previous approaches, FEW allows any learning algorithm in scikit-learn format~\cite{pedregosa_scikit-learn:_2011} to be used for estimation. FEW has been demonstrated for use in regression with several ML pairings, including Lasso~\cite{tibshirani_regression_1996}, linear and nonlinear support vector regression, DT, and k-nearest neighbors (KNN). Central to its ability to evolve features in a single population is the introduction of $\epsilon$-lexicase survival which produces uncorrelated population behavior. 

The wrapper-based ensemble approach to GP is under-studied and presents new challenges from an evolutionary computation standpoint, namely the need for individuals in the population to complement each other in facilitating the learning of the ML method with which they are paired. Our goal in this paper is to use FEW as a test bed for evaluating the ability of several survival and fitness techniques in this new framework for supervised classification. In addition, whereas previously FEW was demonstrated in side-by-side comparisons with default ML methods, here we more robustly analyze whether FEW can, in general, produce better models than existing ML techniques when hyper-parameter optimization of every method is considered. 

This paper contains four main contributions. First, it presents a much-needed analysis of fitness and survival methods for ensemble-based representation learning with GP, which is currently lacking in the field. Second, it focuses on the classification task, which has not been the focus of previous methods with this GP framework. Third, it presents robust comparisons of FEW to other ML methods, including a previously proposed GP method that also focuses on feature learning. As a final contribution we analyze a biomedical problem for which FEW is able to correctly identify the nonlinear, underlying structure of the data across ML pairings, thereby showing the usefulness of learning readable data representations. 

We pair FEW with several well-known classifiers in our analysis: logistic regression (LR), support vector classification (SVC), KNN, DT and random forests (RF). We present an overview of FEW in Section \ref{s:methods} including a description of several fitness and survival methods that are tested. We review related work more thoroughly in~Section \ref{s:related}, including distinguishing between wrapper and filter approaches as well as single, multiple, and ensemble representations of features in GP. The results of the experiments on FEW and its comparison to other methods is shown in Section \ref{s:results}, with discussion and conclusions following in Section \ref{s:discuss}.

\section{Methods}\label{s:methods}

\begin{figure}
\includegraphics[width=\columnwidth]{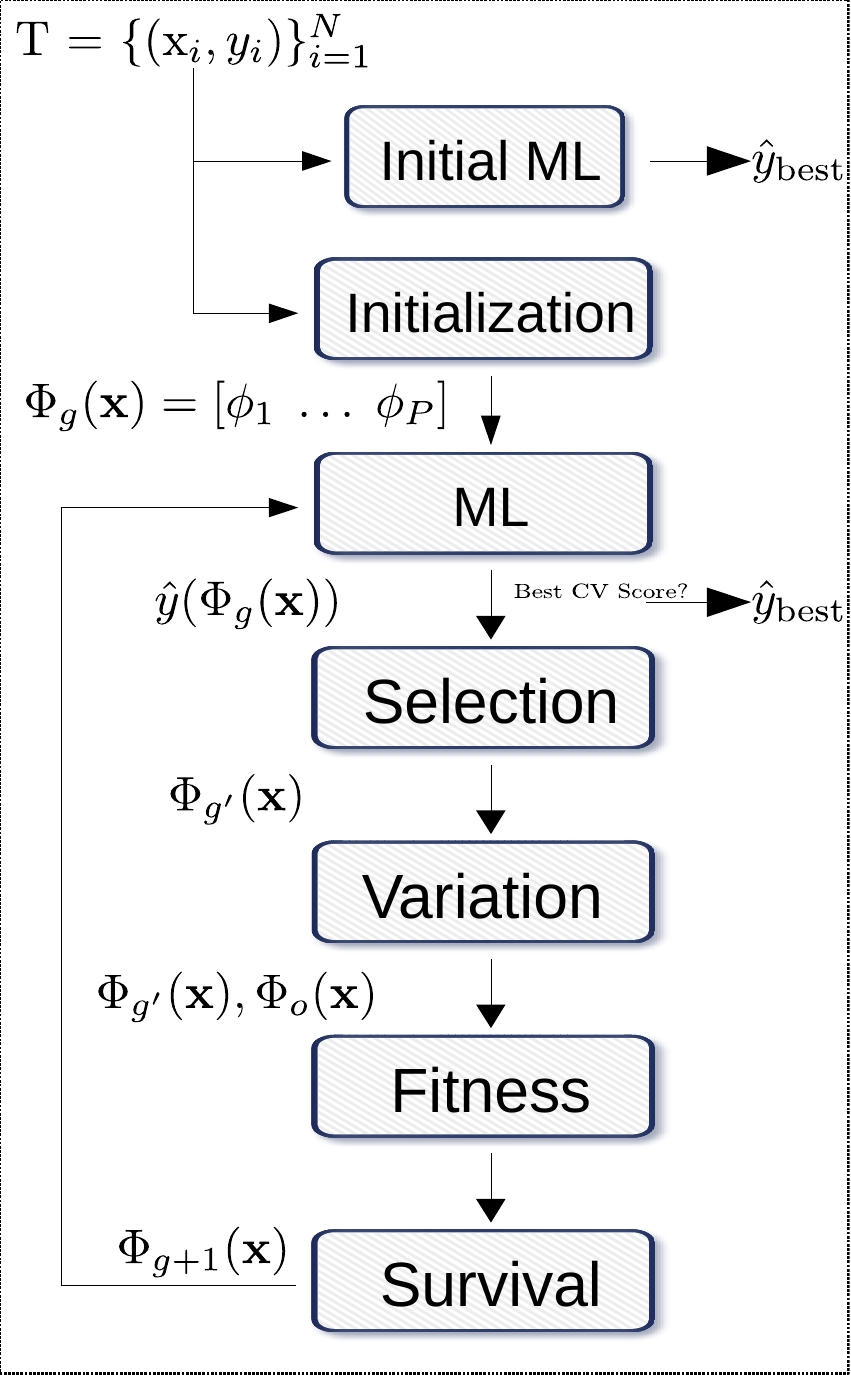}
\caption{\small A diagram showing the main steps in FEW. $\mathbf{\Phi}_{g}$ denotes the starting population; $\mathbf{\Phi}_{g'}$ is the population after selection; $\mathbf{\Phi}_{o}$ is the offspring produced by crossover and mutation; and $\mathbf{\Phi}_{g+1}$ is the new population after conducting survival on $\mathbf{\Phi}_{g'}$,$\mathbf{\Phi}_{o}$.}\label{fig:diagram}
\end{figure}

The components of FEW are summarized in Figure~\ref{fig:diagram}. The learning process begins by fitting the ML method to the original data. FEW maintains an internal validation set to evaluate new models, which guarantees that the returned model will have a cross-validation (CV) fitness at least as good as the initial data representation can produce. FEW then initializes a population of feature transformations, $\mathbf{\Phi}_g(\mathbf{x})$, seeded with the features from the initial ML model with non-zero coefficients. Each generation, a new ML model is trained on $\mathbf{\Phi}_g(\mathbf{x})$ to produce $\hat{y}(\mathbf{\Phi}_g(\mathbf{x}))$. 

The selection step of FEW is the entry point for new information from the ML method about the quality of the current representation. Methods that admit $\ell_1$ regularization (available in the scikit-learn implementations of LR and SVC) or feature importance scores (DT and RF) apply selective pressure to the GP population by eliminating any individuals with a corresponding coefficient or feature importance of zero in the ML model. Feature importance for DT and RF is measured using the Gini importance~\cite{breiman_random_2003}. Thus ML and GP share the feature selection role. After selection, the remaining individuals ($\mathbf{\Phi}_{g'}(\mathbf{x}) \subseteq \mathbf{\Phi}_g(\mathbf{x})$ in Figure~\ref{fig:diagram}) are used to produce $P$ offspring, $\mathbf{\Phi}_{o}(\mathbf{x})$, via sub-tree crossover and point mutation. In this way FEW differs from previous ensemble representation learning approaches~\cite{arnaldo_building_2015,mcconaghy_ffx:_2011} in that it incorporates crossover for variation instead of strict mutation. 

The fitness step (see Section \ref{s:fitness}) evaluates the ability of $\mathbf{\Phi}_{g'}(\mathbf{x})$ and $\mathbf{\Phi}_{o}(\mathbf{x})$ to adequately distinguish between classes in $\mathcal{T}$. The survival step in FEW (see Section \ref{s:survival}) reduces the pool of parents and offspring back to the original size ($P$), and the surviving set of transformations, $\mathbf{\Phi}_{g+1}(\mathbf{x})$, is used at the beginning of the next generation to fit a new ML model.

\subsection{Fitness}\label{s:fitness}
We compare the three fitness metrics (Eqns.~\ref{eq:r2}--\ref{eq:s} below) in our experimental analysis in Section \ref{s:exp_few}. In contrast to traditional GP, the fitness of an engineered feature $\phi(x)$ must measure the individual's ability to separate data between classes rather than its predictive capacity, since $\phi$ is not itself a model. A simple approach to assessing feature quality is to look at the coefficient of determination using 

\begin{equation}\label{eq:r2}
R^2(y,\phi(\mathbf{x})) = 1 - \frac{\sum_i{(y_i - \phi(\mathbf{x}_i))^2}}{\sum_i{(y_i-\bar{y})^2}}
\end{equation}

\noindent For binary classification, $R^2$ seems appropriate, since it only has to capture the correlation of the feature with a change from 0 to 1. However, for multiclass classification, the $R^2$ imposes an additional constraint on the feature by rewarding it for increasing in the direction of the class label values. For certain problems (e.g. one in which the ordering of the class labels corresponds to a degree of risk), this imposed fitness pressure may be warranted, but in the general case we do not want to assume the order of the class labels, nor the relative distance between them in a feature, is meaningful. Instead, we want to reward features that separate samples from different classes and cluster samples within classes. 

Other GP feature construction methods have used the Fisher criterion~\cite{guo_breast_2006,ahmed_multiple_2014} for achieving such a measure. The Fisher criterion assigns fitness of a feature $\phi$ as 
\begin{equation}\label{eq:f}
F = \sum_{i,j \in \mathcal{Y}}{\frac{|\mu_i - \mu_j|}{\sqrt{\sigma_i^2 + \sigma_j^2}}}
\end{equation}

\noindent where $\mu$ is the mean of $\phi(\mathbf{x})$ belonging to a class label, i.e. $\mu_i = \bar{\phi}(\mathbf{x}(|y=i))$, and $\sigma_i$ is the standard deviation. The Fisher criterion gives a measure of the average pairwise separation between, and dispersion within, classes for $\phi$. However, it does not provide fine-grained information about the distance of specific samples in the transformation. In an attempt to extract this information, we include the silhouette score~\cite{rousseeuw_silhouettes:_1987} in our comparisons. Like Eqn.~\ref{eq:f}, the silhouette score assesses feature quality by combining the within-class variance with the distance between neighboring classes. Thus it captures both the tightness of a cluster and its overlap with the nearest cluster. The silhouette score  $s_i$  for a single sample $\mathbf{x}_i$ is defined as 
\begin{align}
a_i &= \frac{1}{|\mathbf{x}^k|}\sum_{\mathbf{x}_j \in \mathbf{x}^k}{||\phi(\mathbf{x}_i) - \phi(\mathbf{x}_j)||_2^2} \nonumber \\
b_i &= \frac{1}{|\mathbf{x}^{k'}|}\sum_{\mathbf{x}_{j'} \in \mathbf{x}^{k'}}{||\phi(\mathbf{x}_i) - \phi(\mathbf{x}_{j'})||_2^2} \nonumber \\
s_i &= \frac{b_i - a_i}{\text{max}(a_i,b_i)} \label{eq:s}
\end{align}
Here, $\mathbf{x}^k = \mathbf{x}(|y=y_i)$ is the set of samples with class label $y_i$, and $\mathbf{x}^{k'}$ is the set of samples in the next nearest class (according centroid distance). Thus Eq.~(\ref{eq:s}) takes into account both the pairwise square distances within a class and the separation of neighboring classes from each other. Here the Euclidean distance metric is used. For aggregate fitness of an engineered feature, the average silhouette score over all samples, $\bar{s} = \frac{1}{N}\sum{s_i}$, is used. 
\subsection{Survival}\label{s:survival}
Unlike typical populations in model-based GP, the surviving individuals in FEW are assessed together in an ML estimation, and therefore benefit from being chosen to work well together. In fact, many ML pairings depend on low co-linearity between features, including LR and SVC. We test four methods for achieving this cooperation: tournament survival (tournaments of size 2), deterministic crowding, $\epsilon$-lexicase survival, and random survival. Tournament survival is agnostic to the population structure when selecting survivors, and simply picks the individual in the tournament with the best fitness to survive. Meanwhile, deterministic crowding and $\epsilon$-lexicase survival are designed to promote feature diversity, which should influence the ability of the population to effectively produce a representation for the ML training step. We include random survival tests to control for the effect of unguided search.  

\textbf{Deterministic crowding}~\cite{mahfoud_niching_1995} is a niching mechanism in which offspring compete only with the parent they are most similar to. We define similarity as the correlation ($R^2$, Eqn.~\ref{eq:r2}) between a child and its offspring. In the case of mutation, there is only one parent, so no similarity comparison is necessary. Although traditionally a steady state algorithm, its implementation here is generational. Children take the place of their parent in the surviving population if and only if they have a better fitness. This algorithm produces niches in the population which should maintain diverse features. 

\textbf{$\epsilon$-lexicase survival} is a new survival technique adapted from $\epsilon$-lexicase selection~\cite{la_cava_epsilon-lexicase_2016} for use in FEW. $\epsilon$-lexicase selection is, in turn, an adaptation of lexicase selection~\cite{spector_assessment_2012, helmuth_solving_2014} for continuous-valued problems. Lexicase selection works by pressuring individuals in the population to solve unique subsets of the training samples (i.e. cases) and shifting selective pressure to cases that are the most difficult in terms of population performance. $\epsilon$-lexicase survival differs from $\epsilon$-lexicase selection in that it removes the individuals selected at each step from the remaining selection pool, and adds them to the survivors for the next generation. Each iteration of $\epsilon$-lexicase survival proceeds as follows:

\smallskip
\noindent{\footnotesize
\begin{tabularx}{\columnwidth}{lX} 
\texttt{GetSurvivors}($\mathcal{\mathbf{\Phi},T}$) 	:						&	\\
\hspace{1em}	$\mathcal{T}' \leftarrow \mathcal{T}$	&	training cases\\
\hspace{1em} $\mathbf{\Phi}_{s} \leftarrow \emptyset$ & survivors\\ 
\hspace{1em}	for each parent selection: & \\
\hspace{1em}\hspace{1em}	$S \leftarrow \mathbf{\Phi} - \mathbf{\Phi}_{s}$	&	initial  pool \\
\hspace{1em}\hspace{1em}	$\epsilon \leftarrow \lambda$($\mathbf{e}_t$) for $t \in \mathcal{T}$	&	get $\epsilon$ for each case \\
\hspace{1em}\hspace{1em}	while $|\mathcal{T}'| >0$ and $|\mathcal{S}|>1$:						&	main loop\\
\hspace{1em}\hspace{1em}\hspace{1em}	\texttt{case} $\leftarrow$ random choice from $\mathcal{T'}$ &	pick a case\\
\hspace{1em}\hspace{1em}\hspace{1em}	\texttt{elite} $\leftarrow$ best fitness in $\mathcal{S}$ on \texttt{case} 	&	determine elite \\
\hspace{1em}\hspace{1em}\hspace{1em}	$\mathcal{S} \leftarrow n \in \mathcal{S}$ if fitness($n$) $\leq$ \texttt{elite}$+\epsilon_{\texttt{case}}$	& reduce pool\\
\hspace{1em}\hspace{1em}\hspace{1em}	$\mathcal{T'} \leftarrow \mathcal{T'} - $ \texttt{case}				&	reduce cases\\
\hspace{1em}\hspace{1em} $\mathbf{\Phi}_{s} \leftarrow  \mathbf{\Phi}_{s} \;\cup$ random choice from $\mathcal{S}$															&	pick survivor  \\
\hspace{1em} return $\mathbf{\Phi}_s$ & return survivors\\ 
\end{tabularx}
}

In the routine above, $\lambda(\mathbf{e}_t) \in \mathbb{R}^P$ is the median absolute deviation of the fitnesses on case $t \in \mathcal{T}$ across the population. 
  

  
\section{Related Work}\label{s:related}
Feature construction has received considerable attention in GP, with implementations falling into single feature, multiple feature and ensemble categories. Single feature representations attempt to evolve a single solution that is an engineered feature as in~\cite{muharram_evolutionary_2005,guo_breast_2006}. Multiple feature representations encode a candidate {\it set} of feature transformations in each individual~\cite{krawiec_genetic_2002,smith_genetic_2005,silva_multiclass_2015,la_cava_william_genetic_2017}, such that each individual is a multi-output estimate of $\mathbf{\Phi}$. In this case, a separate ML model is trained on the outputs of each program, and the resulting output is used to assign fitness to each individual. Ensembles are a more recent approach~\cite{mcconaghy_ffx:_2011,de_melo_kaizen_2014, arnaldo_building_2015,la_cava_general_2017} designed to reduce the computational complexity of fitting a model to each individual. Ensemble approaches instead fit a single ML model to the output of the entire population. This ensemble-like approach treats each individual in the population as single features $\phi$, and treats the ensemble output of the population as $\mathbf{\Phi}$. Among these ensemble methods, FEW shares the most in common with evolutionary feature synthesis (EFS)~\cite{arnaldo_building_2015} in that it uses the more successful wrapper-based approach~\cite{krawiec_genetic_2002, smith_genetic_2005} and incorporates feature selection information from the ML routine. Unlike FEW, EFS pairs exclusively with Lasso~\cite{tibshirani_regression_1996}, uses three population partitions, and does not incorporate crossover between individuals. FEW is motivated by the hypotheses that 1) the ML pairing is best treated like a hyper-parameter of the method, and 2) that existing diversity-preserving selection methods can be successfully adapted to the purposes of ensemble-based feature survival. As a final note, previous work does not often consider the effect of tuning the proposed algorithm or the ML approaches to which is compared, which is a vital step in algorithm comparisons~\cite{caruana_empirical_2006} and in the application of ML to real-world problems. 

\section{Experimental Setup}\label{s:exp}
We conduct two separate sets of experiments. The first set described in Section \ref{s:exp_few} is designed to compare the fitness and survival methods for FEW in combination with different ML methods and hyper-parameters. We use the results the first experiment to choose the fitness and survival method for FEW in the second set of experiments. The second set of experiments, described in Section \ref{s:exp_bench}, is a benchmark comparison of FEW to several ML methods on a larger set of classification problems. All the datasets used in the comparison are freely available via the Penn Machine Learning Benchmark repository\footnote{\url{https://github.com/EpistasisLab/penn-ml-benchmarks}}. 
\subsection{FEW comparisons}\label{s:exp_few}
We tune the choice of fitness and survival methods by performing an experimental analysis of FEW on the tuning problems in Table~\ref{tbl:data} using the parameters listed in Table~\ref{tbl:tuning}. 
\begin{table}
\small
\centering
\caption{\small Tested settings for survival and fitness study.}\label{tbl:tuning}
\begin{tabularx}{\columnwidth}{l R{0.7\columnwidth}}
Setting	&	Values \\ \hline
Population size		&	10, 50, 100	\\
Max depth	&	2,3	\\
Fitness	&	R2, silhouette \\
Survival	&	tournament, deterministic crowding, $\epsilon$-lexicase \\
ML	& LR, DT, KNN \\ \bottomrule
\end{tabularx}
\end{table}

\subsection{Comparison to other methods}\label{s:exp_bench}
We evaluate FEW's performance in comparison to six other ML approaches: Gaussian na\"{i}ve Bayes (NB), LR, KNN, SVC, RF, and M4GP~\cite{la_cava_william_genetic_2017}, a multi-feature GP method derived from~\cite{silva_multiclass_2015} that couples a multi-feature representation with a nearest centroid classifier~\cite{tibshirani_diagnosis_2002}. For more information on the implementations of NB, LR, KNN, SVC, and RF, refer to~\cite{pedregosa_scikit-learn:_2011}. These methods are evaluated on 20 classification problems that vary in numbers of classes, samples and features, as seen in Table~\ref{tbl:main_exp}. To ensure robust comparisons, we include hyper-parameter optimization in the training phase for each method. To do so, we do a grid search of the hyper-parameters of each method (shown in Table~\ref{tbl:main_exp}), using 5-fold cross-validation on the training set to choose the final parameters. The model with the best average cross validation accuracy on the training set is evaluated on the test set. This process is repeated for 30 shuffled, 50/50 train/test splits of the data. In an attempt to control for the different possible hyper-parameter combinations between the methods, we limited each grid search to a maximum of 100 combinations of hyper-parameter settings during training. 

The hyper-parameters considered for FEW (see Table~\ref{tbl:main_exp}) include the population size, the ML method. expressed as a function of the number of features in the data, the output type of the features (float or bool), and max feature depth. Floating point outputs use the operator set $\{+$, $-$, $*$, $/$, $sin$, $cos$, $exp$, $log$, $\sqrt()$, $()^2$, $()^3\}$ and boolean outputs add $\{$\texttt{AND}, \texttt{OR}, \texttt{XOR}, $!$, $==$, $>$, $\geq$, $<$, $\leq\}$. It is important to note that the tuning of the ML method is not considered when paired with FEW. As a result, this experiment compares the relative effects of learning a representation for a default ML method to tuning the hyper-parameters of those methods. 
 
\begin{table}
\scriptsize
\caption{\small Experimental setup for the method comparisons. The hyper-parameters that were searched are shown on the right.  }\label{tbl:main_exp}
\begin{tabularx}{\columnwidth}{X R{0.55\columnwidth}} \toprule
Method& hyper-parameters \\ \midrule
FEW & Population (0.25$d$,$\dots$,3$d$); ML (LR, KNN, RF, SVM); output type (bool, float); max depth (2,3)\\
\rowcolor{Gray}
M4GP & Population size (250, 500, 1000); generations (50,100,500,1000); selection method (tournament, lexicase); max length (10, 25, 50, 100) \\
Gaussian Na\"{i}ve Bayes & none \\
\rowcolor{Gray}
Logistic Regression & Regularization coefficient (0.001,...,100); penalty ($\ell_1$,$\ell_2$,elastic net); epochs (5,10) \\
Support Vector Classifier & Regularization coefficient (0.01,...,100,`auto'); $\gamma$ (0.01, 10, 1000, `auto'); kernel (linear, sigmoid, radial basis function)\\
\rowcolor{Gray}
Random Forest Classifier & No. estimators (10, 100, 1000); minimum weight fraction for leaf (0.0, 0.25, 0.5); max features ($sqrt$, $log_2$, None); splitting criterion (entropy, gini)\\
K-Nearest Neighbor Classifier & K (1,...,50); weights (uniform, distance) \\ \bottomrule
\end{tabularx} 
\end{table}

\begin{table}
\scriptsize
\centering
\caption{\small Classification data sets used in this paper for tuning (top) and comparison to other methods (bottom). GMT stands for GAMETES data sets, which are named according to number of epistatic loci (w), number of attributes (a), noise fraction (n), and heterogeneity fraction (h).}\label{tbl:data}
\begin{tabularx}{\columnwidth}{X r r r} \midrule
Dataset	&	Classes	&	Samples	&	Features\\
\multicolumn{4}{c}{Tuning Problems} \\
\rowcolor{Gray}
auto	&	5	&	202	&	25\\
calendarDOW	&	5	&	399	&	32\\
\rowcolor{Gray}
corral	&	2	&	160	&	6\\
new thyroid	&	3	&	215	&	5\\
\midrule
\multicolumn{4}{c}{Benchmark Problems} \\
\rowcolor{Gray}
analcatdata authorship	&	4	&	841	&	70\\
analcatdata cyyoung8092	&	2	&	97	&	10\\
\rowcolor{Gray}
coil2000	&	2	&	9822	&	85\\
GMT 2w-1000a-0.4h	&	2	&	1600	&	1000\\
\rowcolor{Gray}
GMT 2w-20a-0.4h	&	2	&	1600	&	20\\
german	&	2	&	1000	&	20\\
\rowcolor{Gray}
Hill Valley with noise	&	2	&	1212	&	100\\
Hill Valley without noise	&	2	&	1212	&	100\\
\rowcolor{Gray}
magic	&	2	&	19020	&	10\\
mfeat fourier	&	10	&	2000	&	76\\
\rowcolor{Gray}
mfeat pixel	&	10	&	2000	&	240\\
molecular biology promoters	&	2	&	106	&	58\\
\rowcolor{Gray}
monk2	&	2	&	601	&	6\\
optdigits	&	10	&	5620	&	64\\
\rowcolor{Gray}
parity5+5	&	2	&	1124	&	10\\
schizo	&	2	&	340	&	14\\
\rowcolor{Gray}
texture	&	11	&	5500	&	40\\
vowel	&	11	&	990	&	13\\
\rowcolor{Gray}
xd6	&	2	&	973	&	9\\
yeast	&	9	&	1479	&	8\\
\bottomrule
\end{tabularx}
\end{table} 

\section{Results}\label{s:results}

\begin{figure}
\includegraphics[width=0.95\columnwidth]{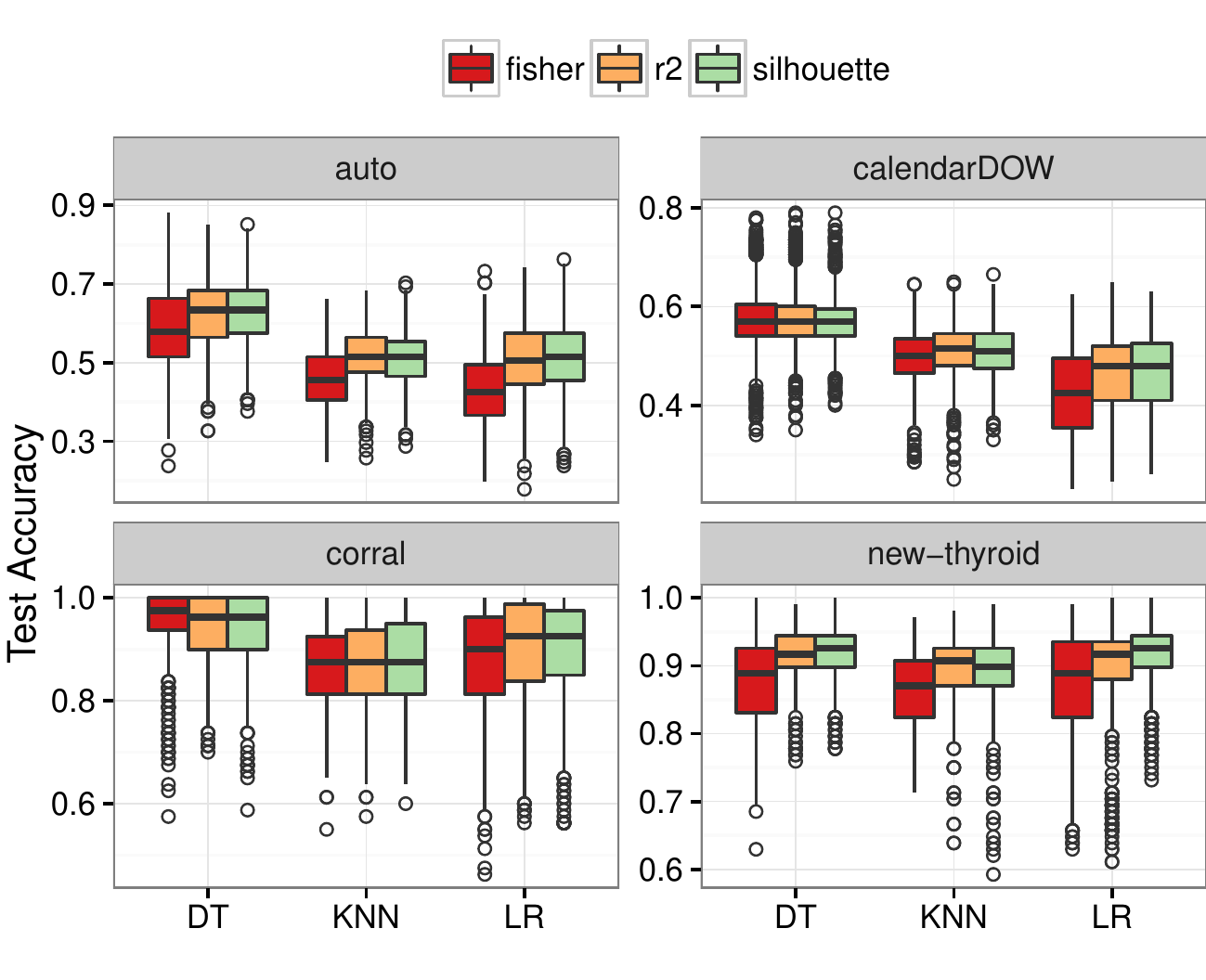}
\caption{\small A comparison of fitness definitions on the tuning data sets. Each subplot presents a different data set; the x-axis corresponds to the paired learner (DT, LR, KNN) and the boxplots represent the accuracy scores obtained with silhouette score (Eqn.\ref{eq:s}) or R$^2$ (Eqn.~\ref{eq:r2}).}\label{fig:tune_fit}
\end{figure}

\begin{figure}
\includegraphics[width=0.95\columnwidth]{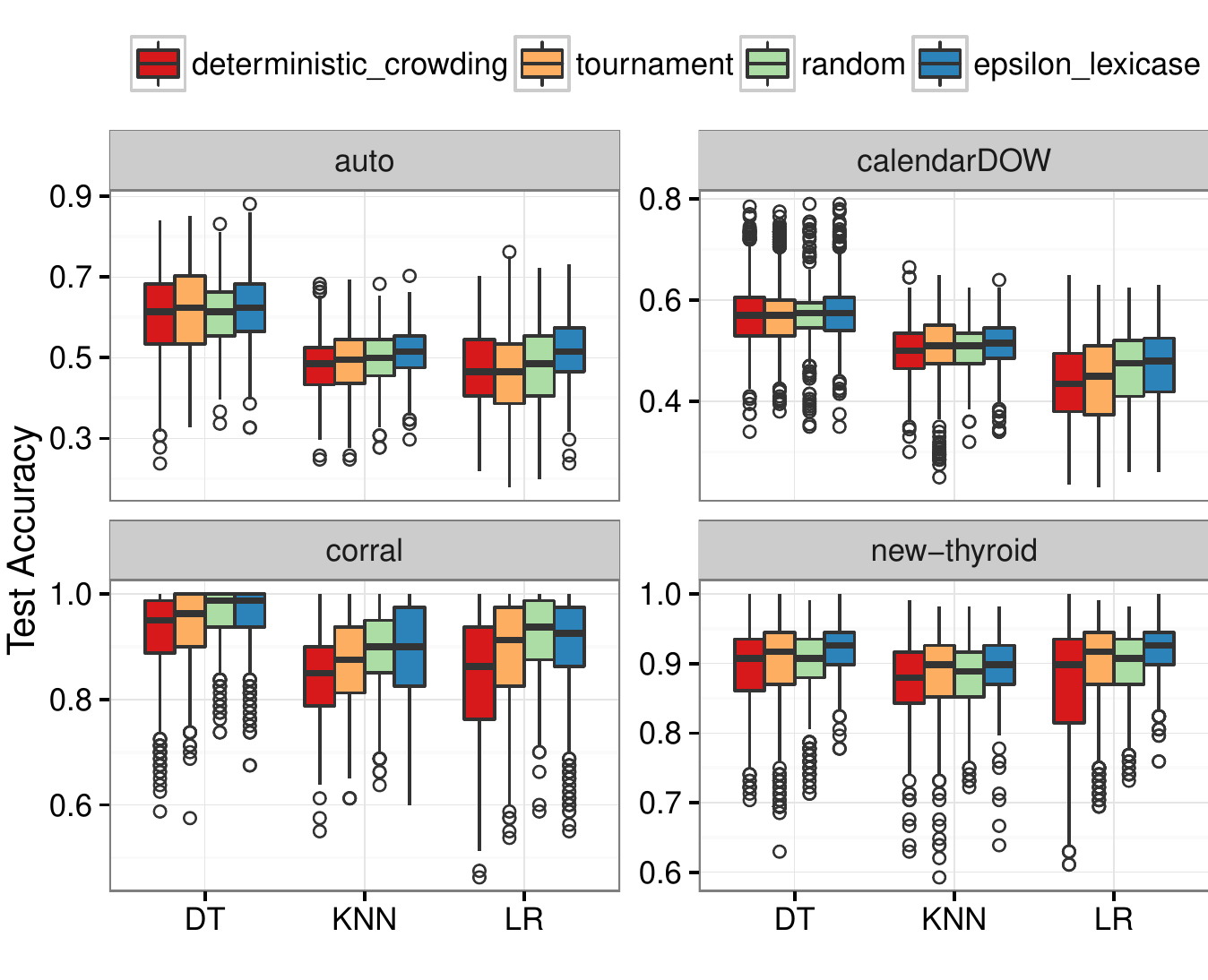}
\caption{\small A comparison of survival algorithms on the tuning data sets. Each subplot presents a different data set; the x-axis corresponds to the paired learner (DT, LR, KNN) and the boxplots represent the accuracy scores using different survival methods.}\label{fig:tune_sel}
\end{figure}

The fitness and survival methods are compared on the tuning datasets in Figures~\ref{fig:tune_fit} and ~\ref{fig:tune_sel}, respectively. The fitness metric comparisons yield unexpected results. The Fisher criterion is outperformed by both R$^2$ and the silhouette score in 3 out of 4 problems ($p<$ 4.8e-7). Surprisingly we find that the silhouette score does not outperform R$^2$ as a fitness metric either; across problems and ML pairings, there is no significant difference in performance aside from new-thyroid. This is surprising given our hypothesis in Section \ref{s:fitness} that the class label assumptions implicit in the R$^2$ would make it less suited to classification with multiple labels. According to this evidence in conjunction with the lower complexity of $R^2$, we opt to use $R^2$ as the fitness criterion for the benchmark comparison.  

We find that $\epsilon$-lexicase survival produces more accurate classifiers than deterministic crowding, tournament and random survival across problems and ML pairings. It is significantly correlated with higher test accuracy according to a t-test ($p<$ 2e-16) and significantly outperforms tournament ($p<$ 0.002) and deterministic crowding ($p<$ 2.4e-7) according to all pairwise Wilcoxon tests, correcting for multiple comparisons. $\epsilon$-lexicase survival also outperforms random survival on auto ($p=$ 4.4e-8) and new-thyroid ($p<$ 2e-16), and ties it on the other two problems (for calendarDOW, $p=$0.094). Random survival performs strongly compared to tournament and deterministic crowding survival, outperforming those methods on 3 out of 4 problems. The results motivate our use of $\epsilon$-lexicase survival in the benchmark comparison. 
 
\begin{figure*}
\includegraphics[width=2\columnwidth]{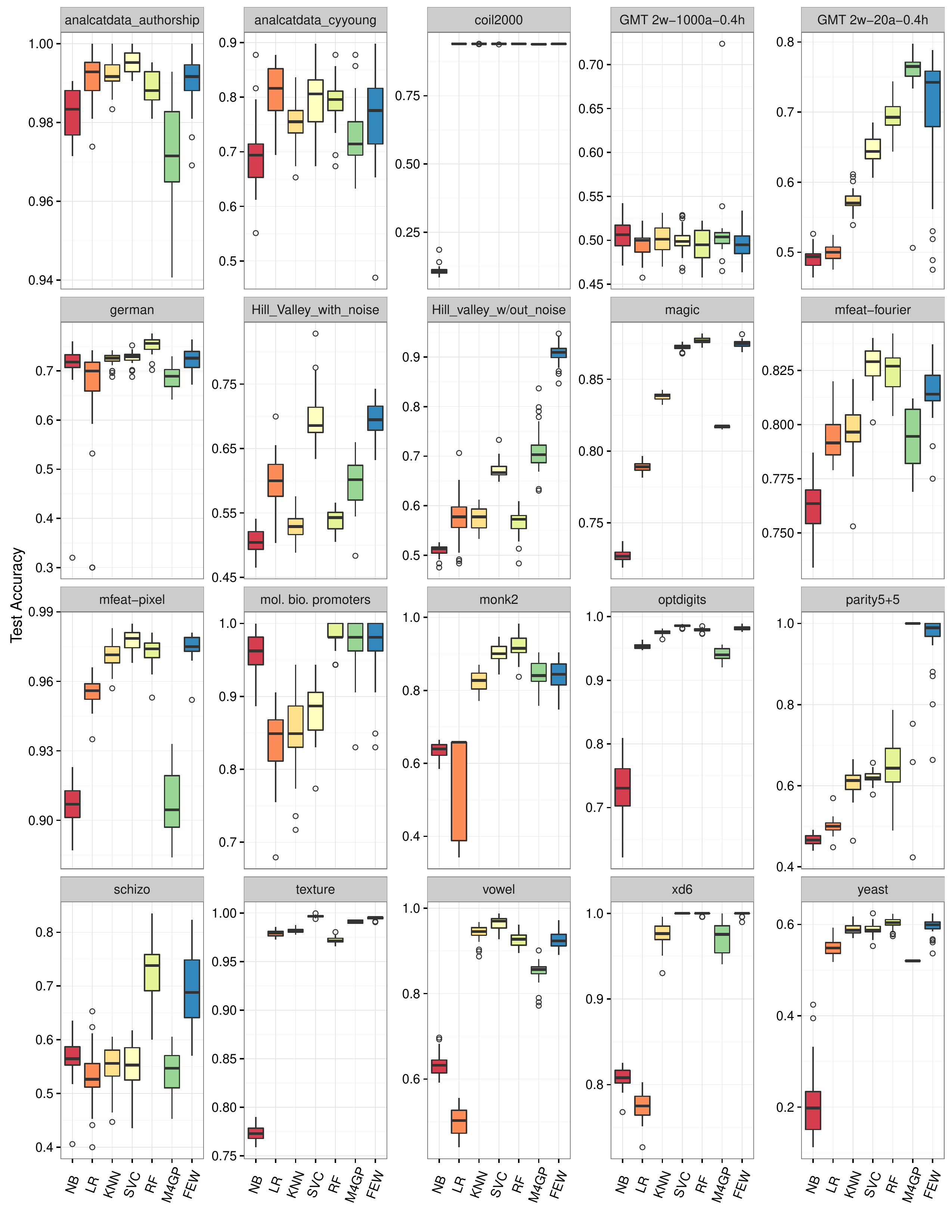}
\caption{\small Comparison of test set accuracy for various methods on the benchmark problems.}\label{fig:all}
\end{figure*}

\begin{figure}
\includegraphics[width=0.8\columnwidth]{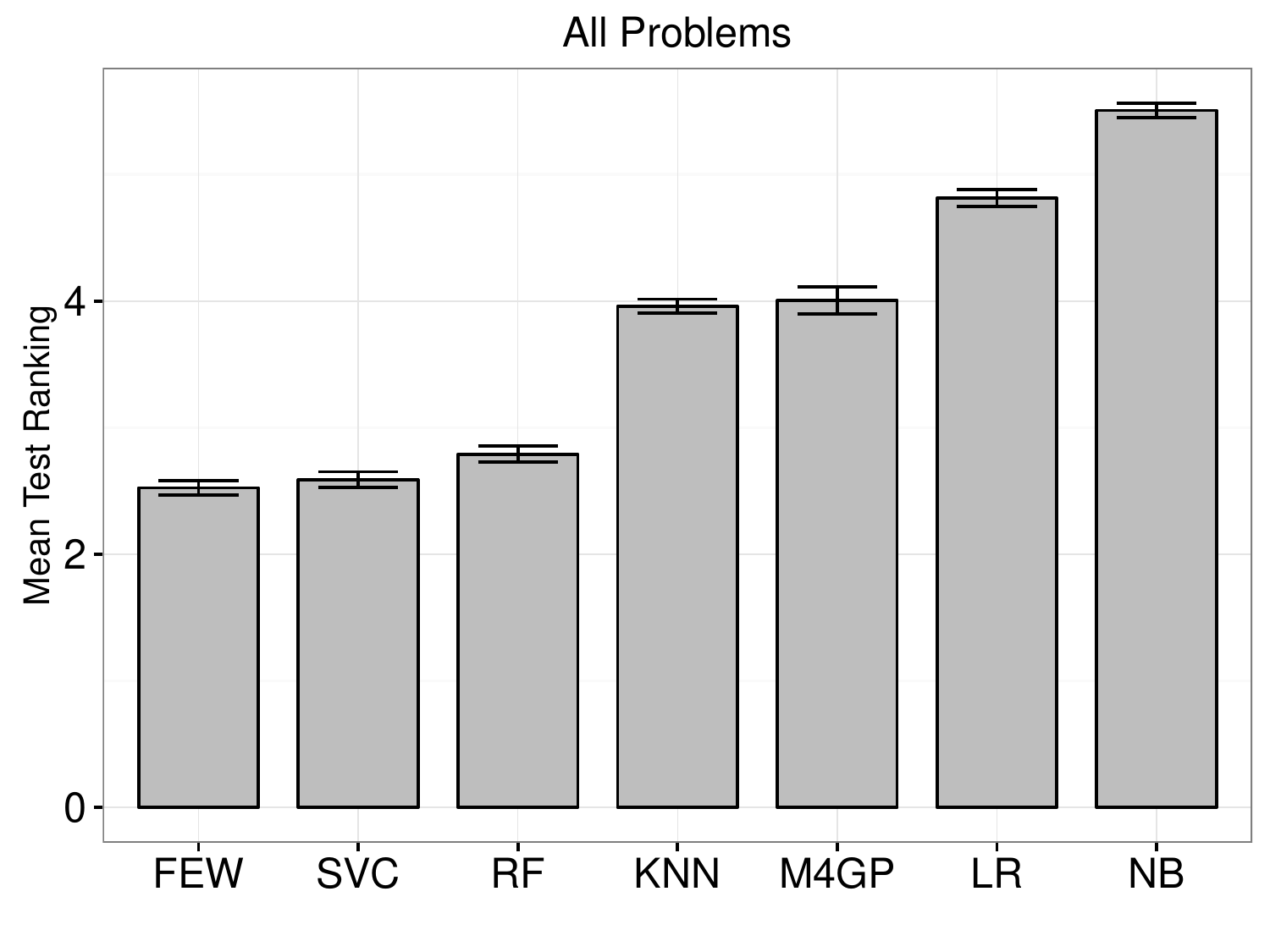}
\caption{\small Ranking of methods over all of the benchmark problems. Bars indicate the standard error.}\label{fig:rank}
\end{figure}

The test set accuracies of the 7 method comparisons on the benchmark datasets are shown in boxplot form in Figure~\ref{fig:all} and the mean rankings are summarized in Figure~\ref{fig:rank}. Across problems, performance varies, generally with RF, SVC, M4GP or FEW producing the highest test accuracy. Whereas FEW generally does well on the problems for which M4GP excels, FEW also does well in cases where M4GP underperforms, which is likely due to FEW's ability to tune the ML method with which it is paired. Three problems stand out for being particularly amenable to feature engineering: GMT 2w-20a-0.4h, Hill\_Valley\_without\_noise, and parity5+5. These three problems are well-known for containing strong interactions between features, which helps explain the observed increase in performance from FEW. In terms of mean rankings across problems, FEW generates the best classifiers among the methods tested, followed closely by SVC and RF. A Friedman test of the rankings with post-hoc analysis reveals RF, SVC, and FEW significantly outperform NB and LR across all problems ($p<$0.039). 

\begin{figure}
\includegraphics[width=0.8\columnwidth]{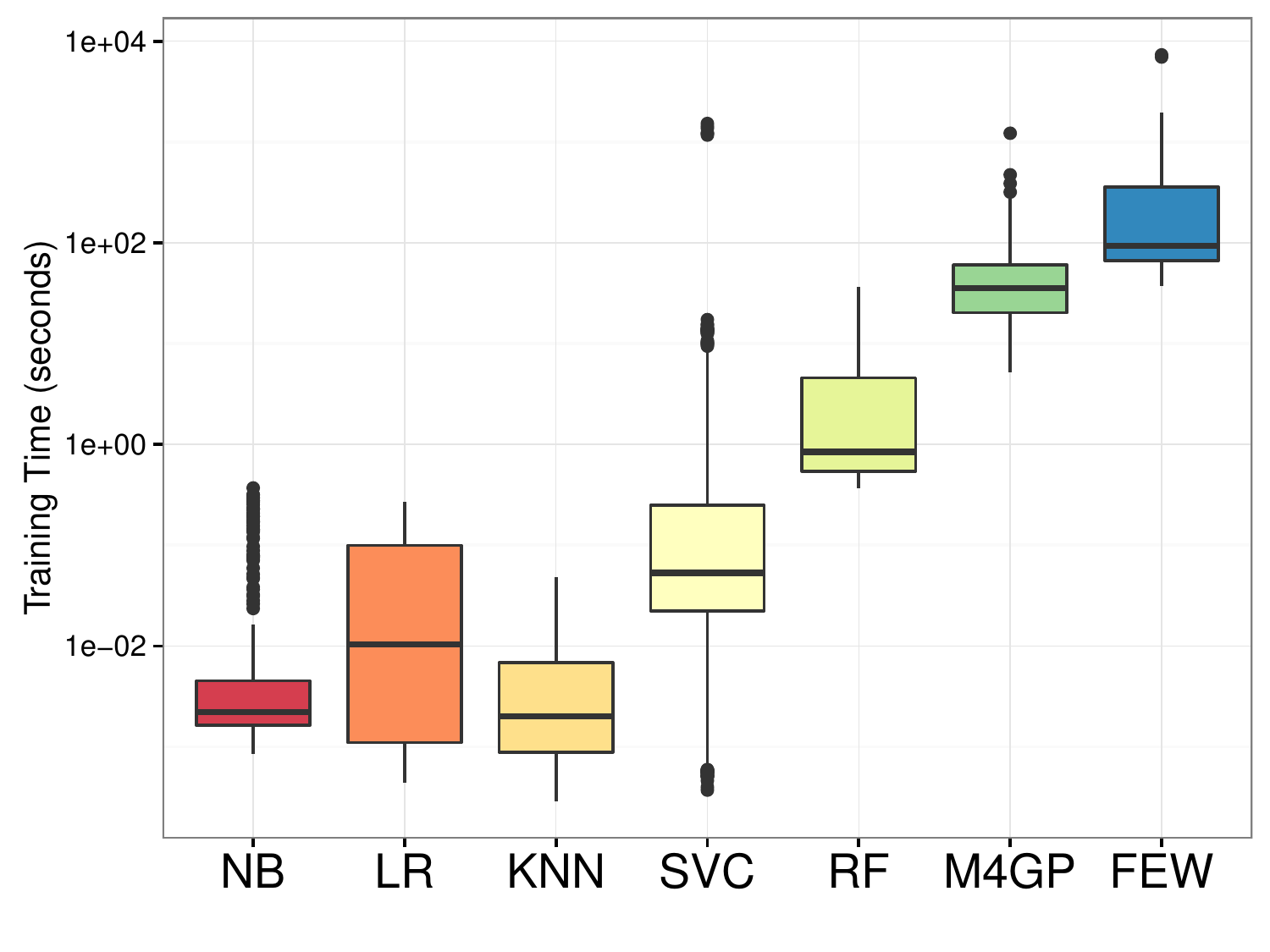}
\caption{\small Training time of methods over all of the benchmark problems.}\label{fig:time}
\end{figure}

As expected, the computation time of FEW is higher than other ML methods (see Figure~\ref{fig:time}) due to its wrapper-based approach. The quicker performance of M4GP may be explained by its c++ implementation compared to FEW's Python implementation, as well as M4GP's use of a consistently fast ML pairing.

We show models generated with single runs of FEW on GMT 2w-20a-0.4h in Table~\ref{tbl:example} using DT and LR. This genetics problem is generated using the GAMETES simulation tool~\cite{urbanowicz_gametes:_2012}. It consists of 20 attributes, 18 of which are noise, and two of which interact epistatically, meaning they must be considered together to infer the correct class (the labels contain noise as well).  The models correctly identify the interaction between features 18 and 19. For this problem FEW's transformation provides the essential knowledge required to solve this problem, whereas the ML approaches simply serve as a discriminant function for processing the information presented via the transformation.




\begin{table}
\centering
\footnotesize
\caption{\small Example solutions to the GMT-2w-20a-40h problem using decision tree and logistic regression pairings. FEW identifies the correct underlying epistatic interaction between features $x_{18}$ and $x_{19}$ in both cases. The final model is either a simple decision tree split in order of the estimated importances or a logistic regression model with four terms.}\label{tbl:example}
\begin{tabularx}{\columnwidth}{Xr} \toprule
\multicolumn{2}{c}{Decision Tree Model} \\
Importance & Feature \\ 
0.899    &    $(x_{19}$ \texttt{XOR} $x_{18})$	\\
0.084    &    $(x_{19}<x_{18})$	\\	
0.017    &    $(\sqrt(|x_8|)>=\cos(x_{18}))$	\\ \midrule
\multicolumn{2}{c}{Logistic Regression Model}\\ 

Coefficient & Feature \\ 
1.992	&	$(x_{19}$ \texttt{XOR} $x_{18})$ \\
1.433	&	$(x_{18}>x_{19})$ \\
0.996	&	$(\exp(x_9)$ \texttt{XOR} $\sqrt(|x_{18}|))$ \\
0.102	&	$(x_4<x_{11})$ \\ \midrule

\end{tabularx}
\begin{tabularx}{\columnwidth}{lrr}
Performance & Decision Tree & Logistic Regression \\ \midrule
Initial ML CV accuracy	 &	0.487	&	0.473	\\
Final model CV accuracy	&	0.763	&	0.803	\\
Test accuracy	&	0.787	&	0.755	\\
Runtime (s)	&	8.2		&	8.2 \\ \bottomrule
\end{tabularx}
\end{table}

\section{Discussion \& Conclusion}\label{s:discuss}
Our results suggest that FEW is a useful technique for supervised classification problems. FEW performs the best on average among the algorithms tested, which include optimized SVM, RF, KNN, M4GP, LR and NB models. This result provides evidence with these ML methods that the data representation can influence algorithm performance as much as, if not more than, the parameter settings of those algorithms. Although it hasn't been tested here, it is likely that including hyper-parameter optimization of the ML methods paired with FEW in the tuning step would show even greater gains in performance over the baseline approach. FEW also performs better than a multiple feature GP approach (M4GP) that uses a fixed ML pairing.  

Despite FEW's runtime in these tests, a complexity analysis suggests it is well-positioned for large datasets in comparison to other feature construction techniques. Whereas techniques like polynomial feature expansion scale poorly with the number of features ($O(d^n)$ for an $n$-degree polynomial) and techniques like kernel transformations scale poorly with the numbers of samples ($O(N^2)$)~\cite{friedman_elements_2001}, FEW scales independently of the features in the dataset, linearly with $N$, and quadratically with the population size. These observations warrant further investigation with large datasets.

\section{Acknowledgements}
This work was supported by the Warren Center for Network and Data Science at the University of Pennsylvania, as well as NIH grants P30-ES013508, AI116794 and LM009012. 



\end{document}